\pdfoutput=1
\documentclass[11pt]{article}
\usepackage{eamt24}

\makeatletter
\newcommand{\@BIBLABEL}{\@emptybiblabel}
\newcommand{\@emptybiblabel}[1]{}
\makeatother
\usepackage[hidelinks]{hyperref}

\usepackage{times}
\usepackage{latexsym}

\usepackage{booktabs, multirow} 
\usepackage{soul}
\usepackage[table]{xcolor} 
\usepackage{changepage,threeparttable} 
\usepackage{graphics}
\usepackage{graphicx}
\usepackage{amsmath,amsfonts,amssymb}
\usepackage{tabularx}
\usepackage{adjustbox}
\usepackage{arydshln}
\usepackage{subcaption}
\usepackage{numprint}

\usepackage[T1]{fontenc}

\usepackage[utf8]{inputenc}

\usepackage{microtype}

\DeclareMathOperator*{\argmax}{arg\,max}

\title{Aligning Neural Machine Translation Models:\\Human Feedback in Training and Inference}

\author{
\textbf{Miguel Moura Ramos}$^{1,2}$ \quad
\textbf{Patrick Fernandes}$^{1,2,3}$  \quad
\textbf{António Farinhas}$^{1,2}$  \quad \\
\textbf{André F. T. Martins}$^{1,2,4}$ \quad    
\vspace{0.2cm}
\\
$^1$Instituto Superior Técnico, Universidade de Lisboa (ELLIS Unit Lisbon)
\\
$^2$Instituto de Telecomunicações\quad 
$^3$Carnegie Mellon University\quad
$^4$Unbabel \\
\small \texttt{miguel.moura.ramos@tecnico.ulisboa.pt}\\
}

\begin{document}
  \maketitle
\begin{abstract}
Reinforcement learning from human feedback (RLHF) is a recent technique to improve the quality of the text generated by a language model, making it closer to what humans would generate.
A core ingredient in RLHF's success in aligning and improving large language models (LLMs) is its \textit{reward model}, trained using human feedback on model outputs. In machine translation (MT), where metrics trained from human annotations can readily be used as reward models, recent methods using \textit{minimum Bayes risk} decoding and reranking have succeeded in improving the final quality of translation.
In this study, we comprehensively explore and compare techniques for integrating quality metrics as reward models into the MT pipeline. 
This includes using the reward model for data filtering, during the training phase through RL, and at inference time by employing reranking techniques, and we assess the effects of combining these in a unified approach.
Our experimental results, conducted across multiple translation tasks, underscore the crucial role of effective data filtering, based on estimated quality, in harnessing the full potential of RL in enhancing MT quality.
Furthermore, our findings demonstrate the effectiveness of combining RL training with reranking techniques, showcasing substantial improvements in translation quality.
\end{abstract}

\section{Introduction}

Neural machine translation (NMT) models \cite{BAHDANAU2015,VASWANI2017} are typically trained with \textit{maximum likelihood estimation} (MLE), maximizing the log-probability of the next word in a translation given the previous words and the source sentence.
While this approach has been effective at training high-quality MT systems, the difference between the training and inference objective can lead to \textit{exposure bias} \cite{BENGIO2015,RANZATO2016,WISEMAN2016}, which hinders the model’s ability to recover from early mistakes. Furthermore, the suitability of model likelihood as a proxy for generation quality has been questioned in machine translation \cite{koehn-knowles-2017-six,ott2018analyzing} and beyond \cite{perez22redteam}. These challenges sparked interest in alternative training and decoding paradigms for MT, such as \textit{reinforcement learning} (RL; \newcite{KREUTZER2018}) or \textit{minimum Bayes risk} decoding (MBR; \newcite{EIKEMA2022}).

More recently, the widespread success of \textit{reinforcement learning from human feedback} \cite{STIENNON2022} has highlighted the importance of a good reward model that approximates well to human preferences for the task at hand. 
While, in general, this requires training a reward model from scratch for the specific problem, in the case of machine translation (MT), the evaluation community has achieved significant progress in developing automatic quality estimation and evaluation metrics \textit{learned} from human quality annotations (e.g. COMET-QE \cite{COMETQE20}, COMET \cite{COMET22}, BLEURT \cite{SELLAM2020}, which can be repurposed as reward models. 
As a consequence, recent research integrating these metrics into the training \cite{REST} or decoding \cite{FERNANDES2022} procedures has had considerable success in improving the quality of translations. However, none of the previous work has systematically compared the effect of integrating metrics at different stages of the MT pipeline or has attempted to combine these techniques in a unified approach.

\begin{figure*}[t]
    \centering
    \includegraphics[scale=0.54]{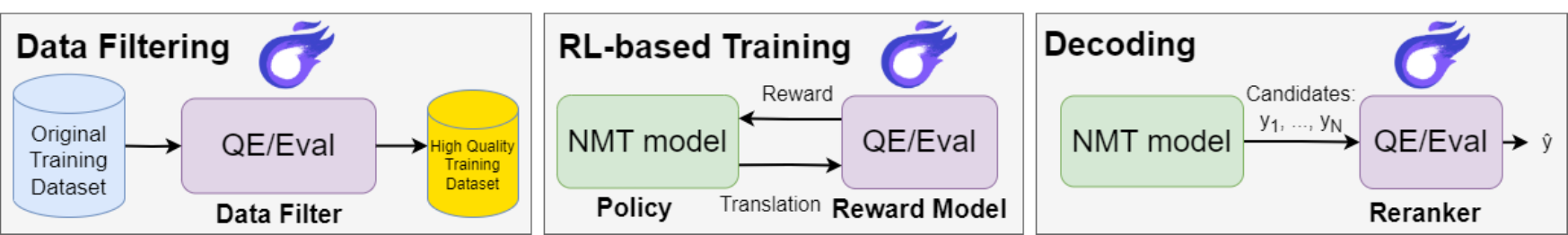}
    \caption{Preference models can have multifaceted roles within the MT pipeline.
    They can serve as effective data filters, refining datasets by incorporating user preferences.
    They can also assume a pivotal role in classic RL training by providing rewards to optimize the MT model performance.
    Finally, they can act as rerankers during the decoding phase, selecting the final translation by maximizing their scores derived from user preferences.
    }
    \label{fig:preference_model}
    \vspace{-4mm}
\end{figure*}

In this work, we perform a comprehensive study on the integration of MT quality metrics into the MT pipeline as reward models. 
As illustrated in Figure \ref{fig:preference_model}, we assess their use at different stages: as a means for data filtering, during the training process through RL, and at inference time by way of reranking techniques. Furthermore, we explore the results of combining these methods. 

We attempt to answer the following research questions:
\begin{itemize}

    \item \textit{Can data filtering based on estimated quality help minimize RL training instability?} 

    \item \textit{Which metrics are more suitable as reward models in RL training? 
    Are reference-free metrics competitive with reference-based ones?}
    
    \item \textit{How does the quality of translations achieved through RL training compare with those produced through reranking approaches? 
    Can these two approaches be effectively combined to further enhance translation quality?} 
\end{itemize}

Our main contributions arise from the research questions mentioned above:
\begin{itemize}

    \item Inspired by \newcite{DATAFILTERING} where they use cross-lingual encoders to score translation representations in an aligned multilingual vector space, we propose an alternative data filtering method that uses COMET-QE \cite{COMETQE20}, a more robust model, to curate a high-quality dataset that empirically helps to \textbf{minimize RL training instability}.
    
    \item We show that neural metrics such as COMET(-QE) \cite{COMET22,COMETQE20} are more suitable than BLEU \cite{PAPINENI2002} for RL training. Contrary to what happens with MBR decoding, RL training results in improved scores across all types of metrics, not only neural ones. In particular, using a reward model based on QE works surprisingly well, possibly paving the way for unsupervised training of NMT systems. 
    
    \item Experiments in EN→DE and EN→FR show that both RL training and reranking techniques enhance translation quality, with RL training often outperforming reranking methods.
    Furthermore, combining RL and MBR decoding results in more consistent improvements across various evaluation metrics.
    
    \item We quantify and discuss the trade-offs in running time at both training and inference, clarifying the efficiency and suitability of each approach.
    
\end{itemize}

\section{Background}

\subsection{Neural Machine Translation}
\label{subsec:Neural Machine Translation}

An NMT model has learnable parameters, $\theta$, to estimate the probability distribution, $p_\theta(y|x)$ over a set of hypotheses $\mathcal{Y}$, conditioned on a source sentence $x$. 
MLE is the training principle of estimating $\theta$, given parallel data, formalized as
\vspace{-3mm}
\begin{equation}
\mathcal{L}(\theta, y_{1:L}) = -\frac{1}{L}\sum_{t=1}^L \log p_\theta(y_t| y_0, . . , y_{t-1}).
\label{MLE}
\end{equation}
NMT systems typically employ \textit{maximum a posteriori} (MAP) decoding to generate translations, 
\vspace{-3mm}
\begin{equation}
    \hat{y}_{\mathrm{MAP}} = \arg \max_{y \in \mathcal{Y}} \log{p_\theta(y | x)},
\label{MAP}
\end{equation}
where algorithms such as greedy decoding or beam search \cite{reddy1977} approximate the most \textit{probable} translation given the source. An alternative approach is to sample translations according to $p_\theta(y|x)$, using techniques such as top-$k$ or nucleus sampling \cite{FAN2018,HOLTZMAN2020}.

In \S\ref{subsec:dp} of this paper, we also consider two distinct reranking approaches \cite{FERNANDES2022}, namely $N$-best reranking and MBR decoding.
While $N$-best reranking selects the candidate translation that maximizes a given (reference-free) metric, MBR decoding ranks candidates using reference-based metrics, maximizing the expected utility (or minimizing the risk).

\subsection{MT Evaluation}
\label{subsec:MT Evaluation}

Human evaluations are the most reliable way to assess the performance of MT systems, but they are time-consuming and costly. 
For that reason, the standard way to evaluate MT is through automatic evaluation metrics, which can be reference-based or quality estimation (QE) metrics.
 
Reference-based metrics compare the generated translation to human-written reference texts.
Lexical reference-based metrics, such as the widely used BLEU \cite{PAPINENI2002}, rely on word overlap and n-gram matching, making them ineffective for translations that have the same meaning but are substantially different from the reference. 
On the other hand, neural metrics, such as COMET \cite{COMET22}, are a recent alternative that relies on neural networks trained on human-annotated data and that leverages contextual embeddings to address semantic similarity.

QE assesses translation quality without human references, being particularly useful in dynamic, data-intensive environments, where references are costly and time-consuming to obtain.
This paper focuses on sentence-level QE as a reward model, providing a single quality assessment for each translation. 
COMET-QE \cite{COMETQE20} is a state-of-the-art reference-free quality estimation metric derived from COMET used to evaluate MT performance. 

Neural reference-based and QE metrics are valuable preference models because they offer a more accurate and contextually-aware measure of translation quality, aligning better with human preferences and judgments \cite{freitag-etal-2022-results}.

\subsection{Reinforcement Learning Training in NMT}
\label{subsec:Reinforcement Learning Training in NMT}
In MT, approaches based on reinforcement learning (RL; \newcite{RL}) cast the problem as a Markov decision process (MDP; \newcite{PUTERMAN1990}), where a source sentence $x = (x_1, ..., x_n)$ is translated into a target sentence $y = (y_1, ..., y_m)$.
Under this perspective, the NMT system can be viewed as the agent with a conditional probability distribution based on its parameters, $p_\theta(y_t |x, y_{<t})$.
The states of the MDP are defined by the target sentence that has already been decoded, $s_t = (y_1, ...., y_{t<m})$, and the action corresponds to the selection of the next word, $y_{t+1}$.
Based on the states and actions, all transitions are deterministic and the reward function, $R$, is provided by the MT evaluation model which returns a quality score for the generated translation $\hat{y}$.
The main purpose of using RL in NMT is to provide learning signals that go beyond a single reference translation, by providing reward signals for arbitrary  translations. 
MLE provides less robust learning signals that are more susceptible to the shortcomings of noisy references. 
However, it is essential to note that if the reward model used relies on reference-based metrics, some vulnerability to noisy references may still persist. 
Accordingly, the goal of RL training is to maximize the expected reward, 
$L_{\mathrm{rl}}(\theta) = \mathbb{E}_{p_\theta(\hat{y} | x)}[R(\hat{y})].$
Commonly used RL training procedures include REINFORCE \cite{WILLIAMS1992}, minimum risk training \cite{OCH2003,SHEN2016}, and proximal policy optimization (PPO; \newcite{PPO}).

\section{Aligning MT with Reward Models}

\subsection{Data Filtering}
\label{subsec:Data Filtering}

The success of fine-tuning NMT models with MLE is highly dependent on the quantity and quality of the training dataset \cite{WANG2018,koehn-knowles-2017-six,khayrallah-koehn-2018-impact}.
This is because accurate references are crucial for computing meaningful learning signals that correctly guide the NMT model towards improved translations \cite{KONG2018}.
Despite its recent successes, RL-based training can be unstable, so using only high-quality data could help mitigate this instability. 
This can be addressed via \textbf{data filtering}, by seeking a good balance between the aggressiveness of filtering and the resulting dataset size: if the original dataset is already small, too much filtering can be detrimental to the performance of NMT systems \cite{ZOPH2016,JIAO2020}.
Furthermore, when looking at the RL scenario, having a sufficiently large training dataset can help guarantee that the NMT model explores a wide range of scenarios for policy improvement.

We apply our data filtering method on the considerably large and noisy WMT datasets \cite{WMT15,WMT16} since they have been reported to have less relevant and uncorrelated sentences that can lead to sub-optimal results when used during training \cite{koehn-etal-2020-findings,WMT22CLEAN}.
We do not perform data filtering to the IWSLT2017 \cite{CETTOLO2012,CETTOLO2017} dataset due to concerns about its limited amount of available data.
Further dataset filtering could potentially result in a too-small training dataset, which is not be desirable for training MT systems.

As illustrated in Figure \ref{fig:preference_model}, to perform the training dataset filtering, we use a filter that reranks the sentence pairs according to quality scores that indicate the correlation and relevance of each sentence and its given reference.
This approach allows us to filter out low-quality sentence pairs, thereby improving the overall quality of the data.
In our approach, we use a robust preference model called COMET-QE \cite{COMETQE20} as the data filter, which combines the use of encoders and a regression model trained on human-annotated data to estimate the quality score of each sentence pair.
This reference-less model is expected to be more accurate in quality score estimation and have a superior alignment with human judgments than just resorting to the currently used cross-lingual encoders which only take into account vector-space mapping similarity \cite{DATAFILTERING}.
Furthermore, COMET-QE seems particularly suitable as our preference model during data filtering, as it is a multilingual reference-free neural-based metric trained on human annotations of translation quality, and therefore can be used to filter by thresholding on predicted quality or on the number of sentences in the training set.
After scoring all sentence pairs, we select the threshold based on the number of high-quality sentence pairs to use as the filtered dataset for RL training.
For that, we apply different thresholds and sizes to the reranked sentences.
We, then, MLE fine-tune our baseline on these subsets and select the subset that gives the overall best-performing model on the dev. set.
These best-performing models serve as baselines for our RL-based training and reranking methods during decoding.

In conclusion, it is worth noting that our data filtering method is, as shown in Figure \ref{fig:preference_model}, one of three methods we cover for employing a preference model in the MT pipeline.
This filtering method can significantly increase the performance of MT systems by introducing feedback in an earlier stage of the pipeline.

\subsection{Training Phase}

The use of RL-based training has the potential to bridge the gap between MLE training objectives, MT evaluation metrics and human-like translations.
However, it faces challenges of instability and inefficiency, especially in gradient variance and reward computation.
As illustrated in Figure \ref{fig:preference_model}, the RL training process is composed of an NMT model that generates translations that are evaluated by the reward model through rewards that represent the quality of the translation.
This reward is used by the policy gradient algorithm to update the NMT model's \textbf{policy}.
To address the problem of gradient variance, we employ PPO \cite{PPO} as our policy gradient algorithm since it is a stable and efficient algorithm that updates the policy parameters in a controlled way with a predetermined proximity bound, avoiding sudden changes that might destabilize the learning.

Reward computation is the most crucial part of this entire process as it guides the NMT model during training. 
Previous work on RL-based NMT systems predominantly used BLEU as the reward function. 
However, BLEU has several limitations, as discussed in \S\ref{subsec:MT Evaluation}. 
To address these shortcomings, we leverage robust preference models during RL training, such as the reference-based COMET \cite{COMET22} and the reference-free COMET-QE \cite{COMETQE20}, as highlighted in Figure \ref{fig:preference_model}. 
Since learning these models is a complex task, we incorporate these pre-trained preference models, which have already been shown to correlate well with human judgments \cite{freitag-etal-2022-results,COMET22,COMETQE20}, to ensure that RL systems can better capture the nuanced preferences of the user by receiving human-like feedback as rewards.
These models assign numerical quality scores to each translation hypothesis based on their desirability, making them similar to utility functions. 
Our study aims to demonstrate that training with RL can generate higher-quality NMT models using neural metrics and investigate the competitiveness of COMET-QE as a reward model.

Another crucial decision was related to the exploitation vs. exploration problem of RL in the context of MT \cite{WU2018}. 
The beam search algorithm generates more accurate translations by exploiting the probability distribution/policy of the NMT model, while sampling aims to explore more diverse candidates.
During generation, we observed that sampling techniques generally led to candidates of lower quality when compared to beam search, according to the preference models used.
Therefore, all RL-based models used beam search during their training and inference. 

\subsection{Decoding Phase}
\label{subsec:dp}
Reranking methods \cite{NGRERANK,ENERGYRERANK,FERNANDES2022,EIKEMA2022} are an alternative to MAP-based decoding that relies on reranking techniques and presupposes access to $N$ candidate translations for each source sentence, generated by the NMT system through methods like beam search or sampling.
The generated candidates are reranked according to their quality given an already determined metric/reward model.

We employ two reranking methods to select a final translation: $N$-best reranking \cite{NGRERANK,ENERGYRERANK} and \textit{minimum Bayes risk} decoding (MBR; \newcite{EIKEMA2022}).

$N$-best reranking \eqref{NBESTRR} employs a reference-free metric, $M_{\textsc{qe}}$, to reorder a set of $N$ candidate translations, denoted as $\bar{\mathcal{Y}}$, and selects the candidate with the highest estimated quality score as the final translation, $\hat{y}_{\textsc{rr}}$,
\vspace{-3mm}
\begin{equation}
    \hat{y}_{\textsc{rr}} = \arg\max_{y \in \bar{\mathcal{Y}}} M_{\textsc{qe}}(y).
\label{NBESTRR}
\end{equation}

Considering the previous equation, and assuming $C_{M_\mathrm{QE}}$ as the computational cost of evaluating a candidate translation with QE metric, $M_\mathrm{QE}$, we obtain the final computational cost of finding the best translation from $N$ candidate translations as $O(N \times C_{M_\mathrm{QE}})$.

MBR decoding, in contrast, relies on a reference-based metric and chooses the candidate that has the highest quality when compared to other possible translations (in expectation). We define $u(y^*, y)$ as the utility function, quantifying the similarity between a hypothesis $y\in\mathcal{Y}$ and a reference $y^* \in \bar{\mathcal{Y}}$. 
In our context, the utility function is represented by either BLEU or COMET.
Therefore, MBR decoding can be  mathematically expressed as
\begin{align}\label{eq:MC-expectation}
    \hat{y}_{\textsc{mbr}} &= \argmax_{y\in\bar{\mathcal{Y}}} ~\underbrace{\mathbb{E}_{Y \sim p_\theta(y \mid x)}[ u(Y, y)]}_{\textstyle \approx ~\frac{1}{N}\sum_{j=1}^{N}u(y^{(j)}, y)},
\end{align}
where in Eq.~\ref{eq:MC-expectation} the expectation is approximated as a Monte Carlo sum using model samples $y^{(1)}, \ldots, y^{(N)} \sim p_\theta(y|x)$.
These samples may be obtained through biased sampling (e.g., nucleus-p or top-k) or beam search.
Knowing that the utility function is a reference-based metric $M_\mathrm{REF}$ with computational cost, $C_{M_\mathrm{REF}}$, and that to find the best translation we need to do pairwise comparisons between hypotheses, we obtain the final computational cost as $O(N^2 \times C_{M_\mathrm{REF}})$.
These reranking methods become particularly effective when $N$ is not excessively large, making the process computationally more manageable.

Preference models capture the preferences of human evaluators and can be used during the decoding stage to influence MT systems, as shown in Figure \ref{fig:preference_model}. 
By doing this, the MT system will prioritize translations that are more aligned with human judgments, therefore reducing the chances of generating severely incorrect translations. 
We believe that incorporating preference models during the decoding stage can lead to even better translation quality, even if the underlying model has already been RL-trained using the same or a different preference model. 
The benefits we expect to see include improved fluency, adequacy, and consistency compared to the respective baselines since our preference models have been trained on annotations that aim to optimize these linguistic aspects. 

\section{Experiments}

\subsection{Setup}

During the training phase, we investigate the advantages of RL training (with and without data filtering \S\ref{subsec:Data Filtering}) for enhancing the performance of NMT systems.
We employ a T5 model\footnote{We leverage the T5-Large model available in Huggingface's \textit{Transformers} framework \cite{HF}.}, pre-trained on the C4 dataset \cite{T5PAPER}.
First, we fine-tune the models using MLE training with Adam \cite{KINGMA2017} as the optimization algorithm, learning rate decay starting from $5 \times 10^{-6}$ and early stopping.
For RL training\footnote{Our RL implementation relies on the Transformer Reinforcement Learning X framework \cite[trlX]{trlx}.}, we use PPO with learning rate set as $2 \times 10^{-5}$, $\gamma$ set as $0.99$, trajectory limit set as $10,000$, beam search size set as $5$ and mini-batch updates were conducted using stochastic gradient descent with a batch size of $32$, gathered over $4$ PPO epochs.
In the inference phase, our emphasis shifts towards reranking techniques and their impact on the performance of NMT systems.
As for the candidate generation method used, early experiments, omitted for relevancy, show that the best configuration is to generate 100 candidates per source sentence and then use sampling with $p = 0.6$ and $k = 300$ to select the best translation.
Consequently, the evaluation encompasses all the baseline and RL-trained models, both with and without $N$-best reranking and MBR decoding. 
These evaluations are conducted across the following datasets:

\begin{itemize}
    \item The small IWSLT2017 datasets \cite{CETTOLO2012,CETTOLO2017} for English to German (EN → DE) and English to French (EN → FR), featuring 215k and 242k training examples, respectively. 
    
    \item The large WMT16 dataset \cite{WMT16} for English to German (EN → DE) with 4.5M training examples.
    
    \item The large WMT15 dataset \cite{WMT15} for English to French (EN → FR) with over 40M training samples.
\end{itemize}

We assess the performance of each NMT system using well-established evaluation metrics, which include BLEU, chrF \cite{POPOVIC2015}, METEOR \cite{BANERGEE2005}, COMET, COMET-QE, and BLEURT. 
Additionally, for certain experiments executed on a single NVIDIA RTX A6000 GPU, we provide wall clock time measurements to offer insights into computational efficiency.

\subsection{Finding the Optimal Quality Subset Size}

\begin{figure*}[t!]
    \centering
    \begin{subfigure}{\textwidth}
        \centering
        \includegraphics[width=0.85\textwidth]{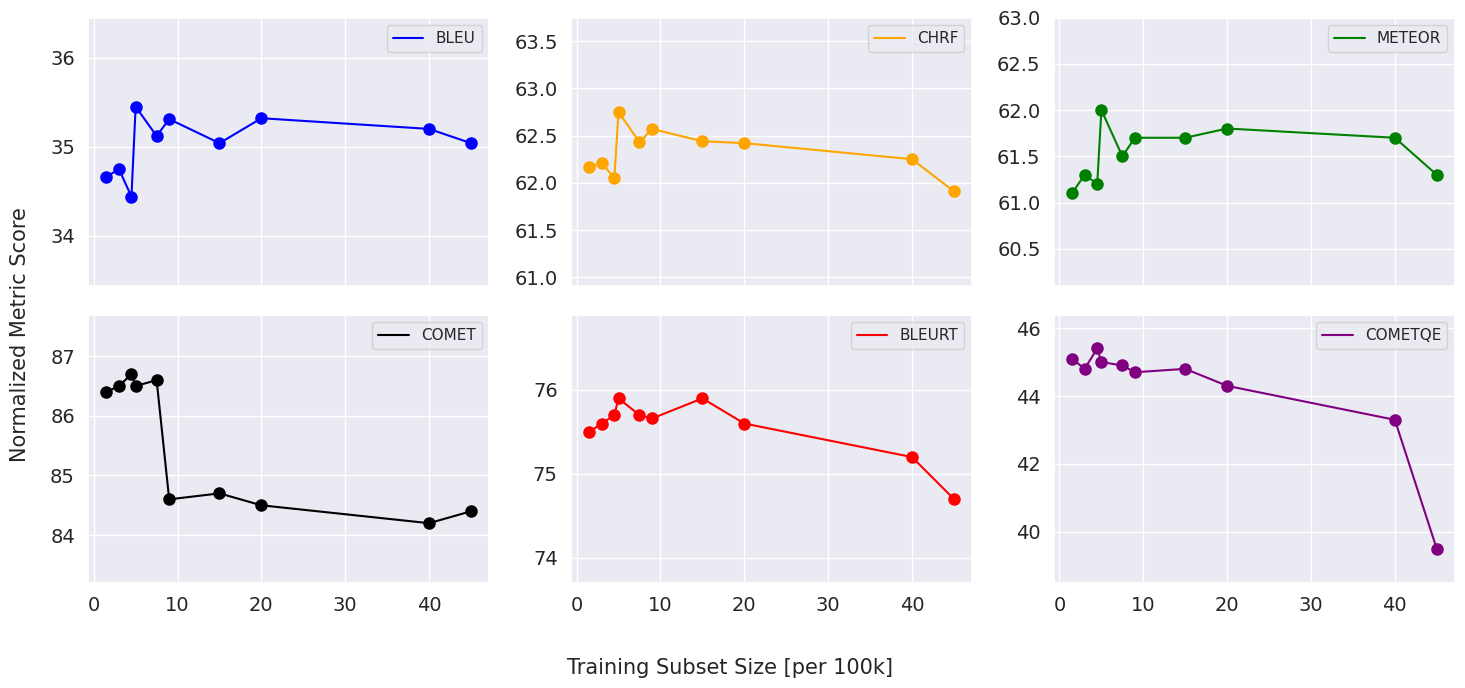}
        \caption{Impact of Data Filtering on WMT16 En→De}
        \label{QSSSEN-DE}
    \end{subfigure}
    \hfill
    \begin{subfigure}{\textwidth}
        \centering
        \includegraphics[width=0.85\textwidth]{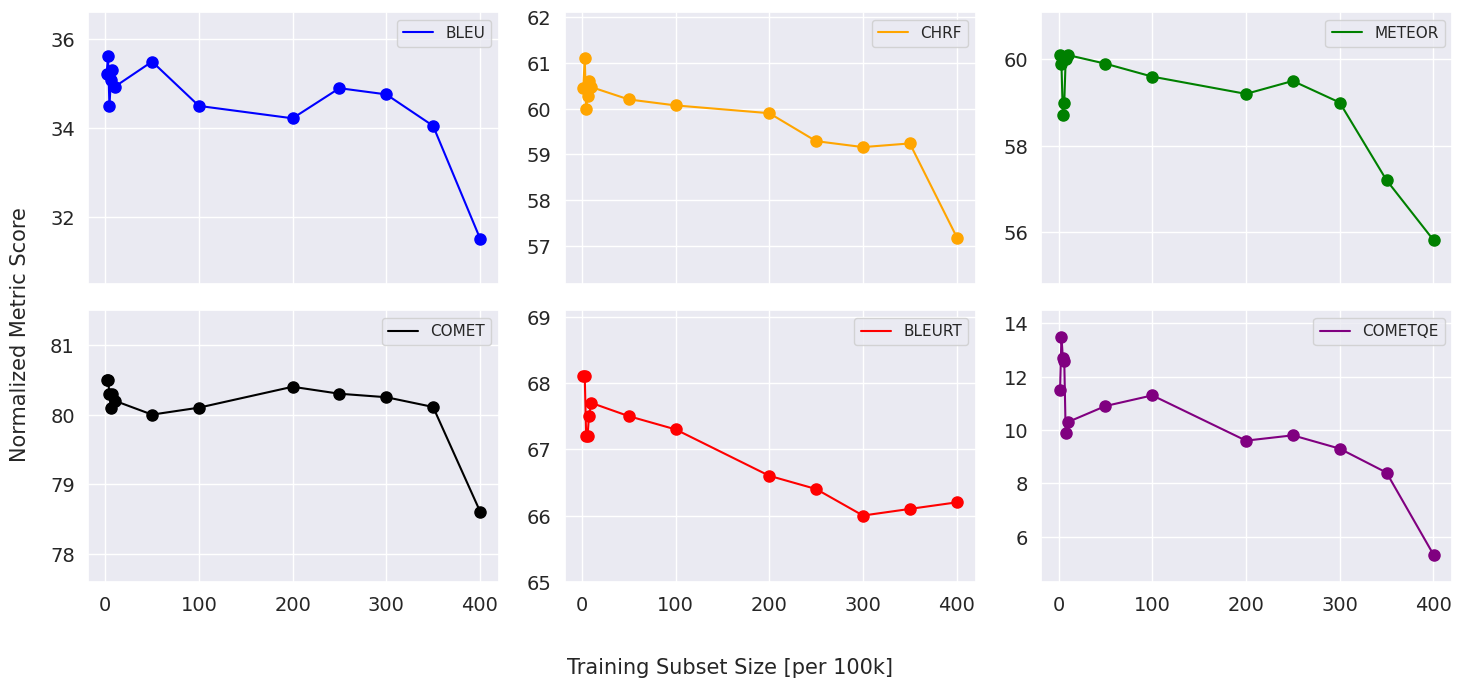}
        \caption{Impact of Data Filtering on WMT15 En→FR}
        \label{QSSSEN-FR}
    \end{subfigure}
    \caption{
    These models were fine-tuned by progressively increasing the size of the high-quality subset, obtained with COMET-QE sentence reranking and denoted in increments of 100,000.}
    \vspace{-1.5em}
\end{figure*}

In this section, we discuss our approach to quality-aware data filtering as a stabilizing strategy (\S\ref{subsec:Data Filtering}), for the WMT datasets.
Figure \ref{QSSSEN-DE} summarizes our findings for the WMT16 EN→DE dataset \cite{WMT16} on the influence of a high-quality subset on translation performance as we vary the subset size, based on various evaluation metrics and COMET-QE sentence filtering.
Across all metrics, a consistent trend emerges: after reaching training sizes of $\numprint{500000}$, there is a notable decline in performance. Particularly, this decline is less prominent for lexical metrics, possibly due to their inherent limitations \cite{freitag-etal-2022-results}.
A similar analysis for WMT15 EN→FR that can be found in Figure \ref{QSSSEN-FR} results in an optimal training size of $\numprint{300000}$ examples.

While the data filtering process has led to remarkable improvements in performance, it is important to note that the effectiveness of this process is dependent on the selected reranking metric. 
Using metrics that are not closely aligned with human judgments can result in poorly correlated and misaligned sentences, which can make the training process more unstable. 
Therefore, it is recommended to use robust QE models, such as COMET-QE.
The more recent COMETKIWI \cite{COMETKIWI2022} model may offer even greater performance improvements.

\subsection{Impact of Quality-aware Data Filtering}

\begin{table*}[!ht]
\centering
\scriptsize
{\renewcommand{\arraystretch}{1.00}
\begin{adjustbox}{scale=0.95}
\begin{tabular}{@{\extracolsep{1pt}}cccccccc}
\toprule
\multicolumn{2}{c}{\textbf{Training Data}} & \multicolumn{3}{c}{\textbf{Lexical Metrics}}&
\multicolumn{3}{c}{\textbf{Neural Metrics}}\\
\cmidrule{0-1} \cmidrule{3-5} \cmidrule{6-8}
SL Data & RL Data & BLEU & ChrF & METEOR & COMET & COMET-QE & BLEURT \\
\midrule
\multicolumn{1}{@{}l}{\raisebox{2ex}{\scriptsize \textbf{MLE}}}\\[-7.5pt]
\rowcolor{orange!30} Original & - &35.04 &61.30 &61.91 &84.40 &39.50 &74.70\\
\rowcolor{orange!30} Random & - &34.43 &61.00 &61.36 &83.90 &39.10 &74.30 \\
\rowcolor{orange!30} XLM-R & - & 33.24 & 60.35 & 60.20 & 84.80 & 41.80 & 72.60 \\
\rowcolor{orange!30} MUSE & - & 35.10 & 61.90 & 62.20 & 85.10 & 40.40 & 74.30 \\
\rowcolor{orange!30} COMET-QE & - &\underline{35.45} &\underline{62.00} &\underline{62.75} &\underline{85.50} &\underline{42.00} &\underline{75.90}\\
\midrule
\multicolumn{1}{@{}l}{\raisebox{2ex}{\scriptsize \textbf{RL w/ BLEU}}}\\[-7.5pt]
\rowcolor{green!20} Original & Original &34.70 &60.90 &61.45 &85.60 &42.20 &74.60 \\
\rowcolor{green!20} Random & Random &34.49 &61.10 &61.49 &85.60 &42.20 &74.40\\
\rowcolor{green!20} XLM-R & XLM-R&33.21 &60.41 &60.10 &85.10 &42.70 &73.10 \\
\rowcolor{green!20} MUSE & MUSE&35.34 &62.10 &62.73 &85.60 &40.80 &74.50\\
\rowcolor{green!20} Original & COMET-QE&35.37 &61.70 &62.04 &85.40 &41.00 &74.20\\
\rowcolor{green!20} COMET-QE & COMET-QE&\underline{35.55} &\underline{62.10} &\underline{62.77} &\underline{\textbf{86.80}} &\underline{\textbf{45.00}} &\underline{\textbf{76.10}} \\
\midrule
\multicolumn{1}{@{}l}{\raisebox{2ex}{\scriptsize \textbf{RL w/ COMET}}}\\[-7.5pt]
\rowcolor{yellow!20} Original & Original&35.05 &61.30 &61.82 &85.60 &41.80 &74.40\\
\rowcolor{yellow!20} Random & Random  &34.96 &61.40 &61.80 &85.60 &41.80 &74.20\\
\rowcolor{yellow!20} XLM-R & XLM-R&33.60 &60.74 &60.40 &85.00 &42.00 &72.90 \\
\rowcolor{yellow!20} MUSE & MUSE &35.18 &61.90 &62.56 &85.50 & 41.90&74.60\\
\rowcolor{yellow!20} Original & COMET-QE &35.58 &61.80 &62.20 &85.70 &41.70 &74.50\\
\rowcolor{yellow!20} COMET-QE & COMET-QE &\underline{35.90} &\underline{62.20} &\underline{63.06} &\underline{86.70} &\underline{44.10} &\underline{75.70}\\
\midrule
\multicolumn{1}{@{}l}{\raisebox{2ex}{\scriptsize \textbf{RL w/ COMET-QE}}}\\[-7.5pt]
\rowcolor{blue!15} Original & Original &34.21 &60.50 &61.10 &85.60 &42.40 &74.80 \\
\rowcolor{blue!15} Random & Random &34.88 &61.30 &61.69 &85.50 &41.80 &74.10\\
\rowcolor{blue!15} XLM-R & XLM-R&33.57 &60.73 &60.40 &85.10 &42.20 &73.20\\
\rowcolor{blue!15} MUSE & MUSE&35.03 &61.90 &62.57 &85.70&41.30 &74.70\\
\rowcolor{blue!15} Original & COMET-QE &35.48 &61.70 &62.10 &85.70 &41.70 &74.50\\
\rowcolor{blue!15} COMET-QE & COMET-QE &\underline{\textbf{35.96}} &\underline{\textbf{62.30}} &\underline{\textbf{63.07}} &\underline{86.70} &\underline{44.70} &\underline{75.90}\\
\bottomrule
\multicolumn{2}{c}{\textbf{Training Data}} & \multicolumn{3}{c}{\textbf{Lexical Metrics}}&
\multicolumn{3}{c}{\textbf{Neural Metrics}}\\
\cmidrule{0-1} \cmidrule{3-5} \cmidrule{6-8}
SL Data & RL Data & BLEU & ChrF & METEOR & COMET & COMET-QE & BLEURT \\
\midrule
\multicolumn{1}{@{}l}{\raisebox{2ex}{\scriptsize \textbf{MLE}}}\\[-7.5pt]
\rowcolor{orange!30} Original & - &31.49 &57.18 &55.80 &78.60 &5.30 &66.20\\
\rowcolor{orange!30} Random & -&31.27 &57.07 & 60.01 & 80.00 &12.80 &65.20\\
\rowcolor{orange!30} XLM-R & - &25.04 &48.78 &  48.60 &77.40 &12.10 &57.10\\
\rowcolor{orange!30} MUSE & - &35.49 & 59.10 & 60.55 & 80.10 & 13.10 & 67.50 \\
\rowcolor{orange!30} COMET-QE & - &\underline{35.62} & \underline{59.90} &\underline{61.11} &\underline{80.50}&\underline{13.50}&\underline{68.10}\\
\midrule
\multicolumn{1}{@{}l}{\raisebox{2ex}{\scriptsize \textbf{RL w/ BLEU}}}\\[-7.5pt]
\rowcolor{green!20} Original & Original &35.47 &59.90 &61.03 &80.20 &16.90 &67.10\\
\rowcolor{green!20} Random & Random &32.75 & 58.10 &60.20 &80.03&14.10 &66.35\\
\rowcolor{green!20} XLM-R & XLM-R&25.78 &49.69 &49.30 &77.70 &13.30 &57.80 \\
\rowcolor{green!20} MUSE & MUSE&35.55 &60.10 &60.56& 81.90&17.10&67.50\\
\rowcolor{green!20} Original & COMET-QE& 35.67 &60.10 &61.01 &81.20 &17.10 &67.30\\
\rowcolor{green!20} COMET-QE & COMET-QE &\underline{36.26} &\underline{60.40} &\underline{61.51} &\underline{82.10} &\underline{17.50}&\underline{67.70}\\
\midrule
\multicolumn{1}{@{}l}{\raisebox{2ex}{\scriptsize \textbf{RL w/ COMET}}}\\[-7.5pt]
\rowcolor{yellow!20} Original & Original&35.50 &59.90 &61.00 &80.40 &16.80 &67.00\\
\rowcolor{yellow!20} Random & Random &34.15 &59.50 &60.93 &80.50 &15.50 &67.10\\
\rowcolor{yellow!20} XLM-R & XLM-R&25.08 &48.84 &48.60 &77.50 &12.40 &57.20 \\
\rowcolor{yellow!20} MUSE & MUSE&36.00 &60.10 &61.20 &80.80&17.00 &67.30\\
\rowcolor{yellow!20} Original & COMET-QE &35.98 &60.00 &61.09 &81.80 &17.10 &67.20\\
\rowcolor{yellow!20} COMET-QE & COMET-QE& \underline{\textbf{36.62}}& \underline{\textbf{60.60}} &\underline{\textbf{61.79}} &\underline{82.20} &\underline{17.40} &\underline{67.60} \\
\midrule
\multicolumn{1}{@{}l}{\raisebox{2ex}{\scriptsize \textbf{RL w/ COMET-QE}}}\\[-7.5pt]
\rowcolor{blue!15} Original & Original &35.50 &60.00 &61.10 &82.20 &17.50 &68.00\\
\rowcolor{blue!15} Random & Random &32.10 &58.30 &60.50 &81.00 &14.40 &66.70\\
\rowcolor{blue!15} XLM-R & XLM-R&24.67 &48.38&48.10 &77.60 &12.60 &56.80 \\
\rowcolor{blue!15} MUSE & MUSE&35.62 &60.45 &59.30 &82.22&17.45 &67.80\\
\rowcolor{blue!15} Original & COMET-QE &35.90 &60.10 &61.22 &82.27 &17.53 &68.02\\
\rowcolor{blue!15} COMET-QE & COMET-QE &\underline{36.25} &\underline{60.50} &\underline{61.58} &\underline{\textbf{82.40}} &\underline{\textbf{17.70}} &\underline{\textbf{68.10}} \\
\bottomrule
\end{tabular}
\end{adjustbox}
\vspace{-0.5em}
\caption{Automatic evaluation metrics for the MLE and RL-trained models on the  WMT16 EN→DE (top) and WMT15 EN-FR (bottom) original datasets, quality subsets obtained from COMET-QE, XLM-R and MUSE and a randomly selected subset. 
The training data used for MLE and RL can be found in the SL and RL Data, respectively.
We experimented with BLEU, COMET and COMET-QE as reward models for the RL training. 
The best overall values are \textbf{bolded} and the best for each specific group are \underline{underlined}.}
\label{tab:NRLvsQSRL}
\vspace{-1.5em}
}
\end{table*}

After obtaining the best configuration for our data filtering process, we experiment with the use of the curated high-quality training subset from COMET-QE and assess its impact on the MLE and RL training performance. 
We compare our filtering method with no filtering by using the original full training dataset, random filtering and cross-lingual embedding similarity filtering using MUSE \cite{MUSE} and XLM-R \cite{XLM-R}. 

Table~\ref{tab:NRLvsQSRL} provides a comprehensive overview of the experimental results using BLEU, COMET and COMET-QE as reward models.
Both MT tasks demonstrate the same tendency when trained using MLE.
COMET-QE and MUSE high-quality subsets have enough reduced noise to provide more stable training,
as evidenced by the overall increase in performance across all metrics compared to the baseline training on the full original dataset. 
Moreover, a randomly selected subset fine-tuned with MLE performs worse or at most not significantly better than the baseline trained on the original dataset, as expected. 
Furthermore, in accordance with our expectations \cite{DATAFILTERING}, XLM-R filtering does not improve training and is actually the worst-performing model.

Regarding RL-based training on both MT tasks, we observe that most RL-trained models outperform their MLE-trained baseline counterparts across various metrics. 
Notably, the best-performing models are the ones that were MLE fine-tuned and then RL-trained on the COMET-QE high-quality subset using both COMET and COMET-QE as reward models.
On top of that, we can see that in some cases RL training solely does not yield significant improvements, but when combined with high-quality training subsets, it results in substantial enhancements and a competitive edge over the normal, random and XLM-R baselines.
Additionally, we see impressive BLEU scores with RL training with COMET(-QE) as reward model.
This finding underscores that optimizing for COMET(-QE) yields superior BLEU scores compared to direct optimization for BLEU. 
This phenomenon is likely attributed to COMET(-QE) providing \textbf{more effective} reward signals during training, thus highlighting the limitations of BLEU.

The excellent performance gains with COMET-QE as a data filter and also as a reward model emphasize the potential of RL-based NMT models trained with a QE reward model (which \textit{does not} require a corpus with references) to outperform other RL-trained models, offering promising opportunities for unsupervised NMT training with monolingual data, especially for low-resource languages, by eliminating the need for reference translations in evaluation and reward signal generation.

In conclusion, we highlight the importance of thoughtful data selection for achieving better translation quality, showing that COMET-QE can consistently outperform the remaining filtering methods.
Furthermore, the top-performing models were RL-trained with \textbf{neural} metrics, showing once again that \textbf{human-aligned} preference models can constantly outperform simpler metrics, such as BLEU.

\subsection{Impact of preference-based MT alignment}

\begin{table*}[t]
\centering
{\renewcommand{\arraystretch}{1.1}
\resizebox{0.98\linewidth}{!}{
\setlength\tabcolsep{4pt}
\begin{tabular}{lccccccccccccc}\toprule
\multirow{2}{*}{{\large\textsc{Model}}} & \multicolumn{6}{c}{WMT16 EN→DE} & \multicolumn{6}{c}{WMT15 EN→FR}
\\ \cmidrule(lr){2-7} \cmidrule(lr){8-14}
& BLEU & METEOR & ChrF & COMET & COMET-QE & BLEURT &
BLEU & METEOR & ChrF & COMET & COMET-QE & BLEURT \\
\midrule
\rowcolor{orange!30}High-Quality Subset Baseline (HQSB)
&35.45 &62.00 &62.75 &85.50 &42.00 &75.90 &35.62 &59.90 &61.11 &80.50 &13.50 &68.10 \\
\hdashline 
\multicolumn{13}{c}{\textit{BLEU}}\\ \hdashline
\rowcolor{green!20}HQSB + RL &\underline{35.55} &62.10 &62.77 &\underline{86.80} &\underline{45.00} &\underline{76.10} &36.26 &60.40 &61.51 &\underline{82.10} &\underline{17.50} &\underline{67.70} \\
\rowcolor{green!20}HQSB + MBR &35.53 &\underline{\textbf{62.30}} &\underline{62.80} &86.70
&44.20 &75.90 &35.73 &60.40 &61.42 &81.60 &15.60 &67.20 \\
\rowcolor{green!20}HQSB + RL + MBR &35.22 &61.90 &62.62 &86.20 &43.10 &75.50 &\underline{\textbf{36.72}} &\underline{\textbf{60.80}} &\underline{\textbf{61.89}} &82.00 &16.30 &67.20 \\
\hdashline \multicolumn{13}{c}{\textit{COMET}}\\ \hdashline
\rowcolor{yellow!20}HQSB + RL &\underline{35.90} &\underline{62.20} &\underline{63.06} &86.70 &44.10 &75.70 &\underline{36.62} &\underline{60.60} &\underline{61.79} &82.20 &17.40 &67.60 \\
\rowcolor{yellow!20}HQSB + MBR &33.58 &60.70 &61.48 &88.00 &\underline{47.90} &76.50 &34.89 &59.60 &60.94 &\underline{\textbf{85.00}} &\underline{27.00} &\underline{69.80} \\
\rowcolor{yellow!20}HQSB+ RL + MBR &34.92 &61.80 &62.84 &\underline{88.10} &47.60 &\underline{76.90} &35.97 &60.20 &61.45 &84.40 &24.50 &69.20 \\
\hdashline\multicolumn{13}{c}{\textit{COMET-QE}}\\ \hdashline
\rowcolor{blue!15} HQSB + RL & \underline{\textbf{35.96}} & \underline{\textbf{62.30}} & \ul{\textbf{63.07}} & 86.70 & 44.70 & 75.90 &\underline{36.25} &\underline{60.50} &\underline{61.58} &82.40 &17.70 &68.10 \\
\rowcolor{blue!15} HQSB + $N$-RR & 31.46 & 58.70 & 60.41 & 87.10 & \underline{\textbf{53.80}} & 75.90 &29.99 &54.80 &56.87 &82.80 &\underline{\textbf{39.10}} &66.20 \\
\rowcolor{blue!15} HQSB + RL + $N$-RR & 32.73 & 59.80 & 61.32 & 87.30 & 53.20 & 76.30 &32.61 &57.40 &58.96 &83.40 &36.10 &67.60 \\
\rowcolor{blue!15} HQSB + $N$-RR + MBR w/ COMET & 33.73 & 60.90 & 61.79 & 88.10 & 49.60 & 76.70 &34.34 &59.40 &60.69 &84.80 &29.40 &69.50 \\
\rowcolor{blue!15} HQSB + RL + MBR w/ COMET & 34.61 & 61.60 & 62.72 & \underline{\textbf{88.20}} & 50.10 & \ul{\textbf{77.20}} &35.47 &59.90 &61.26 &\underline{84.90} &28.80 &\underline{\textbf{70.00}} \\ \toprule
\multirow{2}{*}{{\large\textsc{Model}}} & \multicolumn{6}{c}{IWSLT2017 EN→DE} & \multicolumn{6}{c}{IWSLT2017 EN→FR}
\\ \cmidrule(lr){2-7} \cmidrule(lr){8-14}
& BLEU & METEOR & ChrF & COMET & COMET-QE & BLEURT &
BLEU & METEOR & ChrF & COMET & COMET-QE & BLEURT \\ \midrule
\rowcolor{orange!30} Normal Baseline (NB) &32.75 &62.40 &60.04 &84.80 &38.30 &74.80 &41.47 &68.40 &66.20 &84.40 &21.70 &73.30 \\
\hdashline\multicolumn{13}{c}{\textit{BLEU}}\\ \hdashline
\rowcolor{green!20}NB + RL &\underline{34.48} &\underline{\textbf{62.90}} &\underline{\textbf{60.51}} &\underline{85.20}
&\underline{39.70} &74.40
&\underline{\textbf{44.58}} &68.60 &\underline{66.76} &\underline{85.20} &\underline{24.70} &72.70 \\
\rowcolor{green!20}NB + MBR &33.87 &62.20 &60.05 &85.00 &38.90 &\underline{74.50} &44.08 &\underline{\textbf{68.70}} &66.52 &\underline{85.20} &24.40 &\underline{73.20} \\
\rowcolor{green!20}NB + RL + MBR &34.46 &62.50 &60.22 &85.00 &39.00 &74.10 &44.25 &68.30 &66.50 &85.00 &24.20 &72.40 \\
\hdashline\multicolumn{13}{c}{\textit{COMET}}\\ \hdashline
\rowcolor{yellow!20}NB + RL &\underline{34.17} &\underline{62.20} &59.88 &85.10 &39.30 &74.40 &\underline{44.48} &\underline{\textbf{68.70}} &\underline{66.74} &85.20 &24.60 &72.80 \\
\rowcolor{yellow!20}NB + MBR &33.33 &62.10 &\underline{59.97} &\underline{\textbf{86.70}} &\underline{43.80} &\underline{75.60} &39.04 &65.30 &63.32 &\underline{86.80} &\underline{37.40} &\underline{75.00} \\
\rowcolor{yellow!20}NB + RL + MBR MBR &33.75 &61.90 &59.72 &86.10 &41.80 &74.90 &44.24 &68.50 &66.62 &86.30 &28.30 &73.60\\ \hdashline
\multicolumn{13}{c}{\textit{COMET-QE}}\\ \hdashline
\rowcolor{blue!15}NB + RL 
&\underline{\textbf{34.53}} & \underline{\textbf{62.90}} & \underline{60.49} & 85.30 & 40.00 & 74.70
&\underline{44.56} &\underline{\textbf{68.70}} &\ul{\textbf{66.87}} &85.30 &24.90 &72.90 \\
\rowcolor{blue!15}NB + $N$-RR & 32.31 & 60.70 & 59.06 & 86.40 & \underline{\textbf{50.00}} & 75.60
&42.48 &67.20 &65.38 &86.60 &38.30 &74.00 \\
\rowcolor{blue!15}NB + RL + $N$-RR & 32.98 & 61.50 & 59.48 & 86.40 & 48.70 & 75.40 &43.29 &67.50 &65.90 &86.50 &36.00 &73.70 \\
\rowcolor{blue!15}NB + $N$-RR + MBR w/ COMET & 33.53 & 61.90 & 59.95 & \underline{\textbf{86.70}} & 46.00 & \underline{\textbf{75.80}}
&39.41 &65.40 &63.42 &\underline{\textbf{87.00}} &\underline{\textbf{40.00}} &\underline{\textbf{75.30}} \\
\rowcolor{blue!15}NB + RL + MBR w/ COMET & 34.18 & 62.50 & 60.27 & 86.60 & 43.50 & 75.40 &44.07 &68.20 &66.55 &86.70 &32.50 &74.00 \\ \bottomrule
\end{tabular}
}
\caption{\label{tab:RLvsRanking} Automatic evaluation metrics for the best baseline in each dataset and its variations with RL training, reranking ($N$-RR) and MBR decoding. BLEU, COMET, and COMET-QE serve as reward models in the context of RL training and are subjected to comparison with respect to both reranking strategies employed as the optimization metric (reranker). Best-performing values are \textbf{bolded} and best for each specific group are \underline{underlined}.}
\vspace{-1.5em}
}
\end{table*}

Table \ref{tab:RLvsRanking} presents the performance scores of the best baseline model, across various MT tasks, focusing on the comparison between RL training, reranking methods during inference and the potential synergies between RL training and reranking techniques in improving the translation quality of MT systems.

Our analysis reveals consistent improvements across all evaluation metrics and reward models, with RL training consistently achieving top scores, especially when using COMET-QE as the reward model. \footnote{We also provide additional fine-grained quality analysis in Appendix ~\ref{appendix:AR} to better illustrate and address specific research questions.}
MBR decoding with COMET and \textit{N}-best reranking with COMET-QE outperformed RL training in COMET and COMET-QE metrics but had difficulty improving other evaluation metrics, while RL training exhibited better generalization with slightly less consistent improvements in COMET and COMET-QE scores.
This phenomenon of increased COMET and COMET-QE scores comes at the cost of worse performance according to the other MT evaluation metrics, showing a potential of \textbf{overfitting} effect for these reranking techniques that occur across all datasets.
These findings underscore the potential of neural metrics as reward signals in training and inference, as discussed in \newcite{DEUTSCH2022} and \newcite{freitag-etal-2022-results}.
While combining RL training and MBR decoding occasionally led to top performance, it did not consistently outperform other strategies, making it a method that distributes gains across all evaluation metrics without exceptional generalization as RL training but provides better overall scores than reranking methods alone.

\begin{table*}[htb]
\scriptsize
\centering
{\renewcommand{\arraystretch}{1}
\resizebox{1.00\linewidth}{!}{
\setlength\tabcolsep{4pt}
\begin{tabular}{lcccccccc}\toprule
& \multicolumn{2}{c}{WMT16 EN→DE} & \multicolumn{2}{c}{WMT15 EN→FR} & \multicolumn{2}{c}{IWSLT2017 EN→DE} & \multicolumn{2}{c}{IWSLT2017 EN→FR} \\
\midrule
Method & Training & Inference & Training & Inference & Training & Inference & Training & Inference \\
\rowcolor{orange!30}MLE & 480 & 5 & 373 & 3 & 1020 & 13 & 905 & 16 \\
\rowcolor{green!20}RL & 288 & 5 & 242 & 3 & 354 & 13 & 403 & 16 \\
\rowcolor{yellow!20}MBR & 0 & 212 & 0 & 55 & 0 & 500 & 0 & 660 \\
\rowcolor{blue!15}$N$-RR & 0 & 183 & 0 &  50 & 0 & 455 & 0 & 625 \\
\bottomrule
\end{tabular}
}}
\caption{Wall-clock time values, in minutes, that represent the efficiency of MLE, RL, MBR decoding and $N$-best reranking.
The training was performed on the WMT16 EN→DE and WMT15 EN→FR high-quality subsets and on IWSLT2017 EN→DE and EN→FR entire datasets with $\numprint{500000}$, $\numprint{300000}$, $\numprint{215000}$ and $\numprint{242000}$ sentence pairs, respectively.
The inference was conducted on WMT16 EN→DE, WMT15 EN→FR, IWSLT2017 EN→DE and IWSLT2017 EN→FR official test set partitions with $2999$, $1500$, $8079$ and $8597$ sentence pairs, respectively.
This assessment was done with COMET as the reward model for RL and as a reranker for the reranking methods.
\label{tab:TE}
\vspace{-1.5em}
}
\end{table*}
RL training and MBR decoding in MT exhibit distinct computational efficiency profiles, as shown in Table \ref{tab:TE}.
RL training is computationally demanding but typically entails a one-time, resource-intensive training process (though less resource-intensive than MLE training), involving iterative fine-tuning of NMT models, making it suitable for capturing nuanced quality improvements from the reward models.
In contrast, MBR decoding, focused on optimizing translation during inference, requires recomputation for each input sentence, allowing for computational efficiency when performed infrequently. 
However, it may not fully utilize the capabilities of the NMT model and can be computationally demanding in high-throughput scenarios. 
The choice between RL training and MBR decoding depends on specific MT system requirements, considering computational resources, translation quality objectives, and the need for real-time adaptability.

In summary, the results demonstrate that integrating RL training consistently improves translation quality in both EN→DE and EN→FR tasks across various metrics.
It consistently outperforms the MLE baseline and is superior in lexical metrics scores compared to reranking strategies, which perform well according to COMET and COMET-QE. 
Additionally, most top-performing models incorporate RL training, highlighting its effectiveness in complementing reranking strategies to further improve translation quality.

\section{Related Work}

\paragraph{RL-based NMT.} Extensive research has been conducted on RL algorithms to improve MT. 
Studies by \newcite{WU2018} and \newcite{KIEGELAND2021} have explored the impact of RL training on large-scale translation tasks and demonstrated the effectiveness of policy gradient algorithms in mitigating exposure bias and optimizing beam search in NMT. 
However, both studies were limited to the use of BLEU as a reward model.
Our research differs in that we explore the benefits of employing more robust preference models to improve translation quality.
Additionally, other researchers have made progress in advancing reward-aware training methods.
For instance, \newcite{DONATO2022} introduced a distributed policy gradient algorithm using mean absolute deviation (MAD) for improved training, excelling with BLEU rewards and generalizing well to other metrics. 
Moreover, \newcite{OPENAI} pioneered reinforcement learning from human feedback (RLHF) for a human-based reward model, while \newcite{REST} proposed Reinforced Self-Training (ReST) for more efficient translation quality improvement using offline RL algorithms.

\paragraph{Reranking methods for NMT.} \newcite{SHEN2004} initially introduced the concept of discriminative reranking for Statistical Machine Translation, which was later adopted by \newcite{LEE2021} to train a NMT model through a reranking strategy based on BLEU. 
Extending this concept, 
MBR decoding \cite{KUMAR2004} has regained popularity for candidate generation during decoding, with \newcite{MULLER2021} finding it more robust than MAP decoding, mitigating issues like hallucinations.
Furthermore, \newcite{freitag-etal-2022-high} showed that coupling MBR with BLEURT, a neural metric, enhances human evaluation results when compared to lexical metrics. \newcite{FERNANDES2022} conducted a comprehensive study comparing various reranking strategies, including reranking and MBR decoding, with both reference-based and quality estimation metrics, concluding that these strategies lead to better translations despite the increased computational cost.
In our work, we build on these foundations and show that reranking methods can be coupled with RL training to provide better translation quality to MT systems.
\vspace{-1mm}
\paragraph{Data filtering for NMT.} In their study, \newcite{taghipour-etal-2011-parallel} explored the use of outlier detection techniques to refine parallel corpora for MT.
Meanwhile, \newcite{cui-etal-2013-bilingual} proposed an unsupervised method to clean bilingual data using a random walk algorithm that computes the importance quality score of each sentence pair and selects the higher scores. 
\newcite{xu-koehn-2017-zipporah} presented the Zipporah system, which is designed to efficiently clean noisy web-crawled parallel corpora. 
\newcite{carpuat-etal-2017-detecting} focused on identifying semantic differences between sentence pairs using a cross-lingual textual entailment system. 
\newcite{WANG2018} proposed an online denoising approach for NMT training by using trusted data to help models measure noise in sentence pairs. 
\newcite{LASER} introduced LASER based on a BiLSTM encoder that can handle 93 different languages.
Our work builds on these previous studies as we implement a data filtering method based on COMET-QE, a preference model trained on human preferences. Our approach is similar to that of \newcite{DATAFILTERING} but is significantly more robust as preference models are much more closely aligned to human judgments compared to cross-lingual encoders.

\section{Conclusion}
Our thorough analysis of feedback integration methods underscores the importance of meticulous data curation for enhancing MT reliability and efficiency. 
Our findings demonstrate the consistent improvement in translation quality when employing neural metrics, such as COMET(-QE), during training and/or inference. 
RL training with data filtering stands out as significantly superior to both MLE and reranking methods. 
Additionally, coupling RL training with reranking techniques can further enhance translation quality. 
While computational efficiency remains a concern due to the added overhead of RL and reranking methods on top of MLE-trained models, their adoption should be tailored to specific task and environmental requirements.

\section*{Acknowledgments}

This work was supported by EU's Horizon Europe Research and Innovation Actions (UTTER, contract 101070631), by the project DECOLLAGE (ERC-2022-CoG 101088763), by the Portuguese Recovery and Resilience Plan through project C645008882-00000055 (Center for Responsible AI), and by Fundação para a Ciência e Tecnologia through contract UIDB/50008/2020.

\bibliography{custom}
\bibliographystyle{eamt24}

\clearpage
\appendix

\section{Additional Results}
\label{appendix:AR}

To gain deeper insights into the effectiveness of both training and inference techniques, we also conducted a small fine-grained study evaluating the translation quality of models. Specifically, we compared translations produced by the High-Quality Subset Baseline using three different methods: MBR with COMET, RL training with COMET-QE as a reward model and a hybrid approach combining both.
This complementary evaluation primarily relies on BLEURT, a neural metric highly correlated with human judgments and independent from the used reward models.

\begin{figure}[!h]
    \centering
    \includegraphics[width=0.48\textwidth]{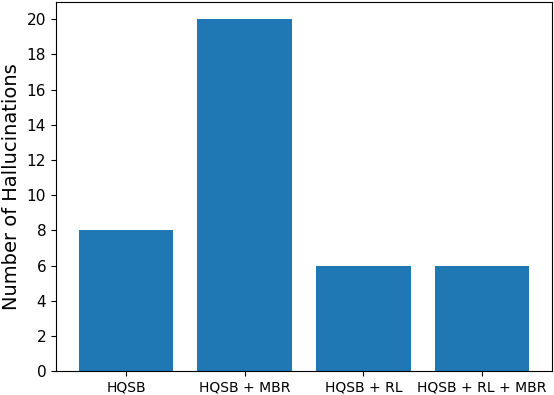}
    \caption{Number of hallucinations on the WMT16 EN→DE test set with $3000$ sentences.}
    \label{fig:hallucinations}
\end{figure}

\begin{figure}[!h]
    \centering
    \includegraphics[width=0.65\textwidth]{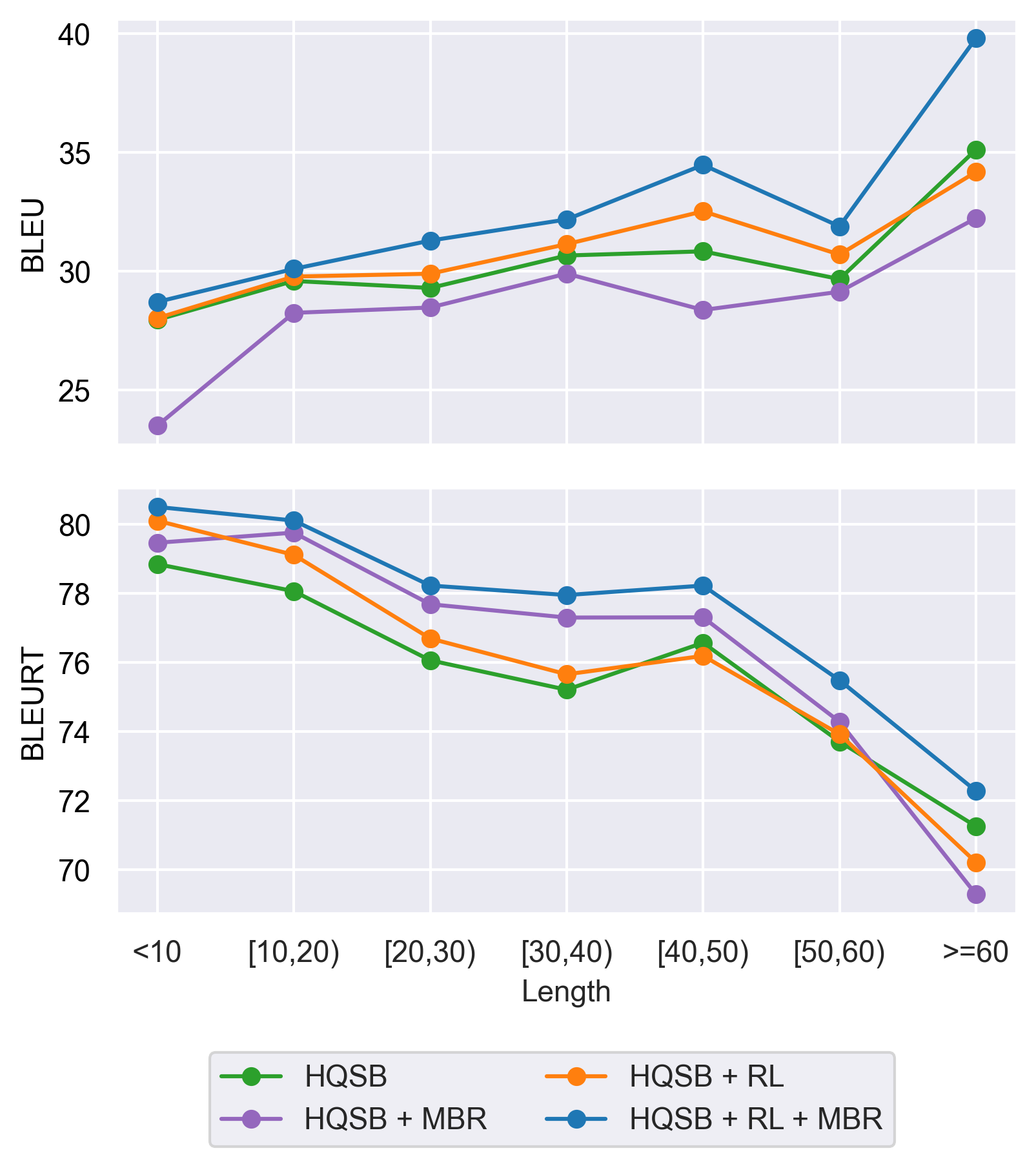}
    \caption{Comparison of BLEU (top) and BLEURT (bottom) scores for WMT16 EN→DE translations across diverse source sentence lengths, highlighting the influence of sentence length on translation quality.}
    \label{fig:length}
\end{figure}

The overall BLEURT scores for these systems can be obtained from Table \ref{tab:RLvsRanking}, with HQSB, HQSB + MBR w/ COMET, HQSB + RL w/ COMET-QE and HQSB + RL w/ COMET-QE + MBR w/ COMET having $75.90$, $76.50$, $75.94$ and $77.20$, respectively.
Figure \ref{fig:length} illustrates a discernible trend: across varying lengths of source sentences, the model trained with RL and employing MBR during inference consistently yields translations of higher quality. Additionally, there is a noticeable decline in translation quality when MBR alone is employed for exceptionally long sentences, a phenomenon seemingly linked to specific hallucinations evident in Figure \ref{fig:hallucinations}. Furthermore, Table \ref{tab:hallucination_table} showcases the most critical examples of hallucinations obtained during this analysis.

\begin{table}[!t]
    \vspace{\baselineskip}
    \centering
    \small
    \renewcommand{\arraystretch}{1.4}
    \begin{tabular}{m{7.3cm}}
\toprule
\textbf{Source:} \textit{Posted by TODAY on Monday, September 14, 2015}
\\
\textbf{Reference:} \textit{Geschrieben von TODAY am Montag, 14. September 2015}
\\
\textbf{MBR Hallucination:} \textit{Posted by TODAY am Montag, 14. September 2015, 14:45 Uhr Posted by TODAY am Montag, September 14, 2015, 14:40 Uhr Posted by TODAY am Montag, September 14, 2015, 14:00 Uhr Posted by TODAY am Montag, September 14, 2015, 14:30 Uhr Posted by TODAY am Montag, September 14, 2015, 14:30 Uhr Posted by TODAY am Montag, September 14, 2015, 14:30 Uhr Posted by TO}
\\
\textbf{RL + MBR Translation:} \textit{Veröffentlicht von TODAY am Montag, 14. September 2015}
\\
\midrule
\textbf{Source:} \textit{Seehofer: "Borders will not be cordoned off"}
\\
\textbf{Reference:} \textit{Seehofer: "Grenzen werden nicht abgeriegelt"}
\\
\textbf{MBR Hallucination:} \textit{Seehofer: "Grenzen werden nicht abgeschottet" Seehofer: "Grenzen werden nicht abgeschottet" Seehofer: "Grenzen werden nicht abgeschottet" Seehofer: "Grenzen werden nicht abgeschottet" Seehofer: "Grenzen werden nicht abgeschottet" Seehofer: "Grenzen werden nicht abgeschottet" Seehofer: "Grenzen werden nicht abgeschottet" Seehofer}
\\
\textbf{RL + MBR Translation:} \textit{Seehofer: "Grenzen werden nicht abgeriegelt"}
\\
\midrule
\textbf{Source:} \textit{Croatia: "We are letting the refugees through"}
\\
\textbf{Reference:} \textit{Kroatien: "Wir lassen die Flüchtlinge durch"}
\\
\textbf{MBR Hallucination:} \textit{Kroatien: "Wir lassen die Flüchtlinge durch" "Wir lassen die Flüchtlinge durch" Kroatien: "Wir lassen die Flüchtlinge durch" Kroatien: "Wir lassen die Flüchtlinge durch" Kroatien: "Wir lassen die Flüchtlinge durch" Kroatien: "Wir lassen die Flüchtlinge durch"}
\\
\textbf{RL + MBR Translation:} \textit{Kroatien: "Wir lassen die Flüchtlinge durch"}
\\
\bottomrule
    \end{tabular}
    \caption{Instances of oscillatory hallucinations generated by the HQSB + MBR model.}
    \label{tab:hallucination_table}
\end{table}

\begin{figure}[!h]
    \centering
    \includegraphics[width=0.8\textwidth]{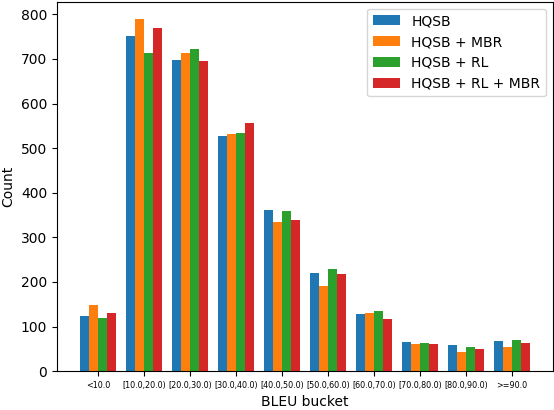}
    \caption{Histograms of sentence BLEU scores for the specific systems on WMT16 EN→DE.}
    \label{fig:bleucount}
\end{figure}

\begin{figure}[!h]
    \centering
    \includegraphics[width=\textwidth]{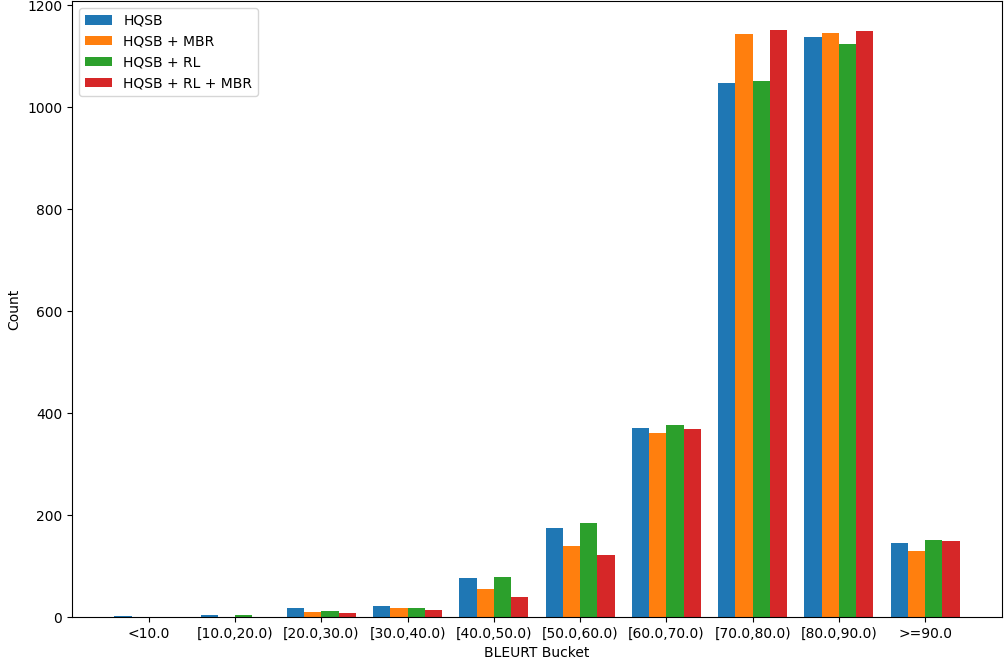}
    \caption{Histograms of sentence BLEURT scores for the specific systems on WMT16 EN→DE.}
    \label{fig:bleurtcount}
\end{figure}

Examining Figures \ref{fig:bleucount} and \ref{fig:bleurtcount}, depicting sentence counts across various ranges of BLEU and BLEURT scores, respectively, reveals the trend that the HQSB + RL + MBR system consistently outperforms the remaining systems across both metrics. 
Once again, the prevalence of low BLEU scores underscores the issue of hallucinations associated with MBR. 
Furthermore, HQSB and HQSB + RL systems are quite competitive but a slight edge must be given to RL in enhancing the performance of the models

\begin{figure}[!h]
    \centering
    \includegraphics[width=\textwidth]{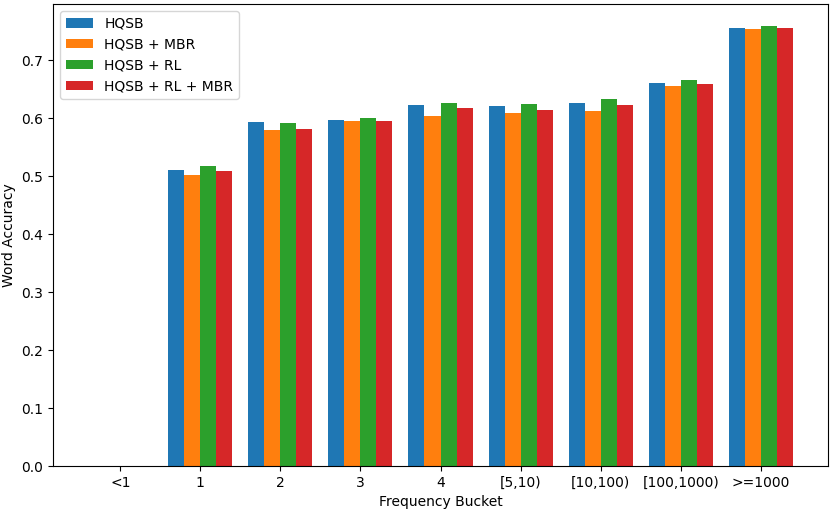}
    \caption{Word F-Measure Bucketed by Frequency for the specific systems on WMT16 EN→DE.}
    \label{fig:waa}
\end{figure}

The bucketed word accuracy analysis aims to evaluate how effectively each system is at generating different types of words. Figure \ref{fig:waa} shows that all four systems demonstrate robustness across all word frequencies but perform significantly better with higher-frequency words. Notably, among these systems, the one integrating reinforcement learning (RL) emerges as the top performer, emphasizing its effectiveness in word generation tasks.

\clearpage
\onecolumn

\end{document}